\pdfoutput=1

\documentclass[11pt]{article}

\usepackage{ACL2023}

\usepackage{times}
\usepackage{latexsym}

\usepackage[T1]{fontenc}

\usepackage[utf8]{inputenc}

\usepackage{microtype}

\usepackage{inconsolata}

\usepackage{enumitem}
\usepackage{graphicx}
\usepackage{booktabs}
\usepackage{multirow}
\usepackage{amsmath}
\usepackage{amsmath}

\renewcommand{\sectionautorefname}{\S\kern-2pt}
\renewcommand{\subsectionautorefname}{\S\kern-2pt}
\renewcommand{\subsubsectionautorefname}{\S\kern-2pt}

\newcommand{\tabuc}[2][c]{\begin{tabular}[#1]{@{}c@{}}#2\end{tabular}}  
 
\newcommand{\tabul}[2][c]{\begin{tabular}[#1]{@{}l@{}}#2\end{tabular}}   

\makeatletter
\newcommand{\myitem}[2][]{%
\item[#2#1]\protected@edef\@currentlabel{#2}%
}
\makeatother

\title{SpaceNLI: Evaluating Natural Language Inference Models\\on Spatial Reasoning}
\title{SpaceNLI: Test Bed for Consistent Inferences in Space}
\title{SpaceNLI: Test Bed for Consistent Natural Language Inferences in Space}
\title{SpaceNLI: Evaluating the Consistency of Predictions In Space}
\title{SpaceNLI: Evaluating the Consistency of Predicting Inferences In Space}

\author{Lasha Abzianidze \hspace{10mm} Joost Zwarts \hspace{10mm} Yoad Winter\\
  Institute for Language Sciences, 
  Utrecht University\\
  Utrecht, the Netherlands\\
  \texttt{\{l.abzianidze, j.zwarts, y.winter\}@uu.nl}
}

\date{}

\begin{document}
\maketitle
\begin{abstract}
While many natural language inference (NLI) datasets target certain semantic phenomena, e.g., negation, tense \& aspect, monotonicity, and presupposition, to the best of our knowledge, there is no NLI dataset that involves diverse types of spatial expressions and reasoning.
We fill this gap by semi-automatically creating an NLI dataset for spatial reasoning, called SpaceNLI.%
\footnote{\url{https://github.com/kovvalsky/SpaceNLI}}
The data samples are automatically generated from a curated set of reasoning patterns (see \autoref{fig:intro}), where the patterns are annotated with inference labels by experts.
We test several SOTA NLI systems on SpaceNLI to gauge the complexity of the dataset and the system's capacity for spatial reasoning.
Moreover, we introduce a \emph{Pattern Accuracy} and argue that it is a more reliable and stricter measure than the accuracy for evaluating a system's performance on pattern-based generated data samples.
Based on the evaluation results we find that the systems obtain moderate results on the spatial NLI problems but lack consistency per inference pattern.
The results also reveal that non-projective spatial inferences (especially due to the ``between'' preposition) are the most challenging ones.
\end{abstract}

\section{Introduction}

Natural language inference (NLI) is a popular task that evaluates NLP systems on text reasoning skills.
In the task, a system has to predict an inference relation from a premise text to a hypothesis sentence/phrase.
Usually, the task is three- or two-way classification, depending on whether in the inference labels of \emph{entailment}, \emph{neutral}, and \emph{contradiction}, the latter two are merged into \emph{non-entailment}.
The task is intended for evaluation of NLP systems on reasoning, however, the systems with competitive results on NLI benchmarks are often exploiting dataset biases (\citealt{tsuchiya-2018-performance,poliak-etal-2018-hypothesis,gururangan-etal-2018-annotation,mccoy-etal-2019-right}, \textit{inter alia}) and their performance suffers from out-of-distribution NLI sample problems \citep{glockner-etal-2018-breaking}.

\begin{figure}[t]
\mbox{\includegraphics[clip, trim=0mm 13mm 44mm 0mm, width=.48\textwidth]{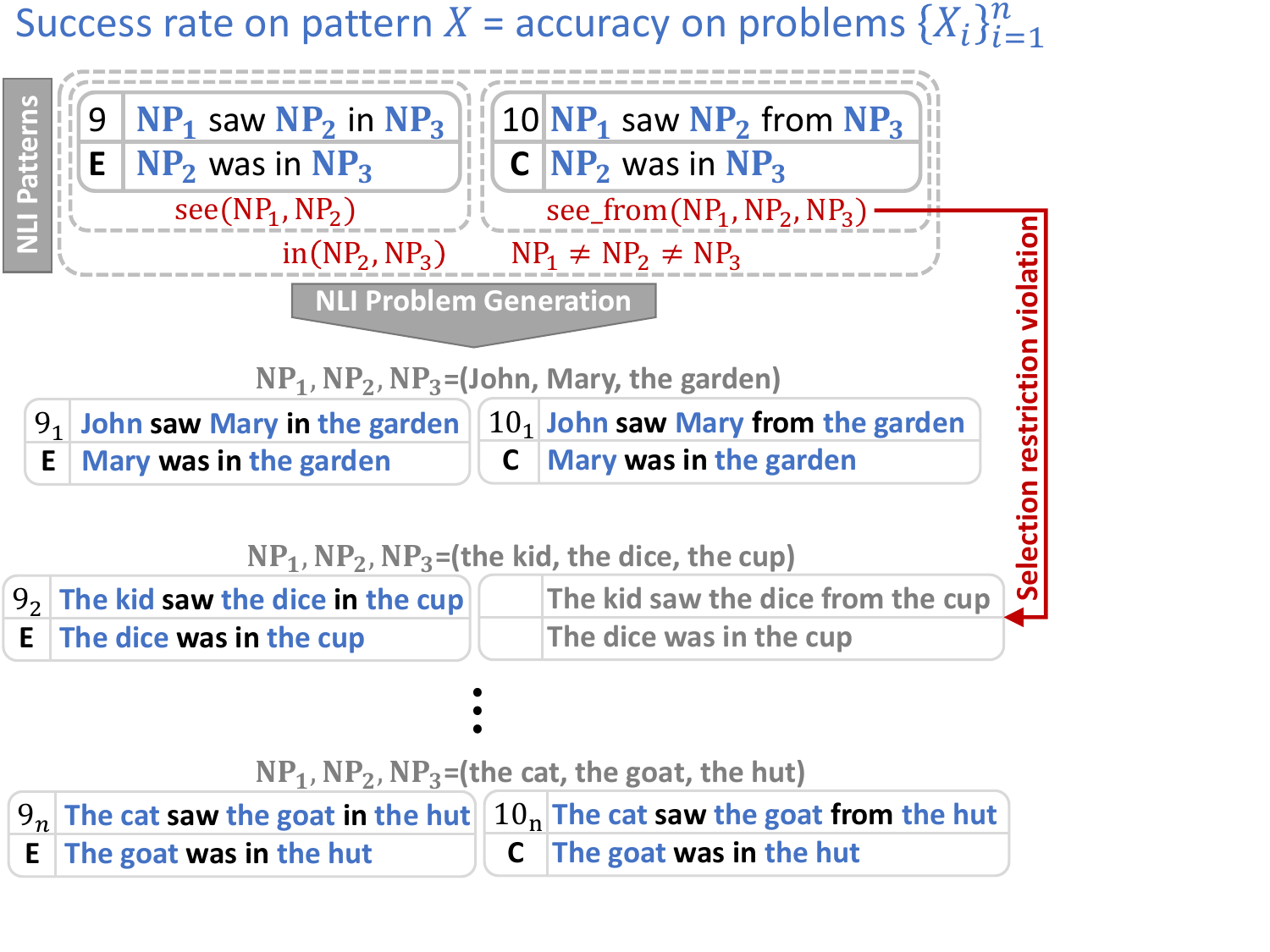}}
\caption{Sampling NLI problem from NLI patterns (with IDs 9 and 10, \textbf{E}ntailment and \textbf{C}ontradiction, respectively). 
The problems are generated by replacing NP placeholders with definite NPs that satisfy pattern-specific selection restrictions.
A system's success rate on a pattern is defined as the accuracy on its corresponding NLI problems.}
\label{fig:intro}
\end{figure}

To better evaluate the reasoning skills of NLI systems, a series of works have been \mbox{(semi-)automatically} or manually creating NLI datasets that specialize in certain semantic phenomena.
While some of these datasets come with a training part, most of them are intended solely for evaluation.
For example, several datasets have been dedicated to monotonicity reasoning \citep{yanaka-etal-2019-help,yanaka-etal-2019-neural,yanaka-etal-2020-neural}, negation was targeted by \citet{hossain-etal-2020-analysis}, the dataset by \citet{kober-etal-2019-temporal} focuses on temporal and aspectual inferences,   \citet{jeretic-etal-2020-natural} semi-automatically generated NLI problems for implicatures and presuppositions.
There are also NLI datasets that cover several semantic phenomena, having a separate section for each of the phenomena (\citealt{fracas,richardson_sem_frag:2020}, \textit{inter alia}).

While spatial reasoning has been included in several multi-modal QA datasets \citep{VQA:2015,suhr-etal-2017-NLVR,Johnson_2017_CLEVR,hudson2019gqa} and in a couple of text-based QA datasets \citep{WestonBCM15,mirzaee-etal-2021-spartqa}, to the best of our knowledge, no NLI dataset has specifically covered it.%
\footnote{Even the FraCaS dataset \citep{fracas,maccartneythesis}, which was curated by linguists and semanticists, doesn't cover spatial semantics within its nine sections.}
This paper fills the gap by semi-automatically creating an NLI dataset for spatial inferences.
First, we collected a diverse set of NLI problems inspired by the inference examples found in the literature on spatial semantics.
Second, the NLI problems were manually converted into NLI patterns (see \autoref{fig:intro}), and finally, we automatically generated a large number of NLI problems from the patterns.

The paper makes two main contributions:
\begin{enumerate}[
topsep=0pt, itemsep=2pt, partopsep=2mm, parsep=0pt]
    \myitem[.]{C1} \label{c1} 
    SpaceNLI: the spatial NLI dataset with diverse types of spatial inferences;
    The inference labels of the generated problems are highly faithful (97\%) to the labels of the corresponding original patterns.  
    \myitem[.]{C2} \label{c2}
    Pattern accuracy and its curve: they measure systems' performance on patterns and the consistency in predictions on samples from the same patterns.
\end{enumerate}

The conducted experiments answer the following research questions: 
\begin{enumerate}[
topsep=0pt, itemsep=2pt, partopsep=2mm, parsep=0pt]
     \myitem[.]{Q1} \label{q1}
    How much spatial reasoning current SOTA NLI systems are capable of?  
    \myitem[.]{A1} \label{a1}
    We found out that the SOTA NLI systems have problems with fine-grained spatial inferences. Their performance drops at least by 24\% compared to their results on common NLI datasets. 
    Moreover, their consistency in predictions is sensitive to irrelevant lexical substitutions. 
    \myitem[.]{Q2} \label{q2}
    What types of spatial inference problems are easy or challenging for the SOTA NLI systems? 
    \myitem[.]{A2} \label{a2}
    The results showed that the non-projective spatial relations are most challenging for the models. 
    This was mainly due to difficulty associated with ``between'' and its frequent occurrence in the evaluation dataset.  
\end{enumerate}

\section{Spatial expressions and inferences}\label{sec:spatial}
\subsection{Types of spatial expressions}

Spatial expressions consist of spatial prepositions and other expressions with spatial information (e.g., \textit{far}, \textit{the left of}, and \textit{in front of}).   
They usually describe a relation between two entities, the \emph{figure} and the \emph{ground}.
The site or path of the figure is the focus of the discussion and is characterized with respect to the ground. 
For example, in ($9_1$) and ($10_1$) from \autoref{fig:intro}, \textit{Mary} is a figure and \textit{garden} a ground.
\textit{John} is also a figure in the premise of ($10_1$).

Spatial expressions are roughly divided into \emph{locative} and \emph{directional} expressions, where locatives can be further classified into \emph{projective} and \emph{non-projective} \citep{Herskovits:86}.
The locative expressions describe static, locative relations between the figure and the ground while directional ones describe a more \emph{dynamic} relation involving a movement and/or path.
An example with a directional preposition is \textit{Cindi walked into the market}.
The spatial expressions in \autoref{fig:intro} are all locative except for \textit{from}, which is directional. 
These locative expressions are non-projective since they require only the spatial location of the figure and the ground.
In contrast, projective locatives additionally require further information from the ground in terms of a~deictic frame of reference (i.e., an orientation structure).
For example, the site of the house is not sufficient to interpret  Mary's location in \textit{Mary is behind the house}, it requires knowledge about the frame of reference of the house, in particular, what counts as a back side of the house. 

\begin{table*}[t]
\centering
\scalebox{.77}{\begin{tabular}{@{}c c l c l@{}}
\textbf{~ID} & \textbf{Class} & \textbf{Premise(s)} & \textbf{L} & \textbf{Hypothesis}  
\\\midrule
15 & Dir  
& John threw the ball into the box.
& E & The ball went into the box.
\\\midrule
16 & Dir  
& John threw the ball at the box.
& N & The ball went into the box.
\\\midrule
31a & Dir  
& Los Angeles is in California. John came from California.
& N & John came from Los Angeles.
\\\midrule
38 & NonP  
& John is in the garden. The garden is in the church.
& E & John is in the church.
\\\midrule
41 & Dir 
& John drove through the tunnel.
& E & John was in the tunnel.
\\\midrule
47a & Dir 
& Cindi walked into the market.
& E & Cindi was outside the market.
\\\midrule
56c & Proj 
& The trash can is to the right of the tree from John.
& C & \small{The tree is to the right of the trash can from John.}
\\\midrule
70 & Proj 
& Mary is between the tree and the house. The tree is behind the house.
& E & Mary is behind the house.
\\\midrule
80 & NonP 
& \footnotesize The cat is between the house and the fence. The cat is between the fence and the tree.
& C & \footnotesize The cat is between the house and the tree.
\\\midrule
99*d & Proj 
& The bucket is above the bowl. The pencil is above the bowl.
& N & The bucket is below the pencil.
\\\midrule
96b & ArgO 
& Mary met John at the party.
& N & Cindi was not at the party.
\\\midrule
100 & NonP 
& The house is far from the school.
& E & The school is far from the house.
\\\midrule
102a & ArgO 
& Mary has taken the cup out of the cabinet.
& C & The cup is in the cabinet.
\\\midrule
102f & ArgO 
& Mary has hidden the cup behind the cabinet.
& E & The cup is not in the cabinet.
\\\midrule
\end{tabular}}
\caption{Examples of the seed NLI problems annotated with spatial inference classes: \textbf{Dir}ectional, \textbf{Proj}ective, \textbf{Non-P}rojective, and \textbf{Arg}ument \textbf{O}rientation. 
Initial letters abbreviate the corresponding inference labels.
}
\label{tab:seed_probs}
\end{table*}

\subsection{Types of spatial inferences}\label{ssec:inf_types}

We characterize spatial inferences depending on the type of spatial expressions licensing them.
An inference might depend on several spatial expressions of a different type, which makes partitioning the inferences challenging, if not impossible.
We define the following classes that represent a coarse-grained partition of spatial inferences. The classes will be later referred to in \autoref{sec:data}.%
\footnote{Licensing contradiction and neutral problems will be assumed from the perspective of a related entailment problem.
For example, we assume that the neutral problem (16) in \autoref{tab:seed_probs} is licensed in the same way as its related entailment (15).
Put differently, one can see (16) as an adversary to (15) and assume that solving (15) requires competence comparable to the one required for solving (16).
}

\paragraph{Argument orientation}
In spatial literature, an argument orientation entailment identifies which argument of the verb is the figure of the spatial expression.
For instance, (9$_1$) in \autoref{fig:intro} show that \textit{Mary} is the figure of the locative PP \textit{in the garden}. 
In its original interpretation, the argument orientation entailment is not restricted to spatial expressions of a particular type.
Here, we restrict the class of argument orientation to the entailment problems (and their neutral and contradiction counterparts) that come close to resolving a PP attachment.
For example, correctly resolving the PP attachment in (9$_1$) boils down to the hypothesis.   
The problems in this class contain a hypothesis with a copula and a predicative spatial PP, where the PP is contrasted to a tightly related PP in the premise(s).
For more examples of the NLI problems in the argument orientation class, see \autoref{tab:seed_probs}. 

\paragraph{Directional}
The directional class contains spatial inferences where directional spatial expressions play the key role. 
Examples of such inferences are given in \autoref{tab:seed_probs}.
Some of these NLI problems pertain to a path-place relation:
(47a) shows that \textit{walking into} infers \textit{being outside};%
\footnote{Since moving along the path is related to the change of the location, sometimes spatial entailments interfere with tense and aspect.     
}
(41) entails \textit{being in the tunnel} from the premise that states that the driving path was through the tunnel.
(31a) combines a part-whole relation with the movement path. 

\paragraph{Projective}
This class contains inferences that hinge on a frame of reference introduced by projective spatial expressions.
In principle, the frame of reference can introduce six directions that can be referred to using the expressions like \textit{front, behind, left, right, above, below, under, on top of}, etc. (see the examples of NLI problems in \autoref{tab:seed_probs}).
The NLI problems that contain \textit{on top of} as only projective spatial expression, and when its projective interpretation is not crucial for the inference, are put in a different class.

\paragraph{Non-projective}
We classify a problem as having non-projective inference if the inference is driven only by non-projective spatial expressions. 
Therefore, an occurrence of non-projective spatial expressions in a problem is necessary but not sufficient for assigning the problem to this class, e.g., see directional problems (31a) and (41).
NLI problems that depend on spatial expressions with the semantics of order and proximity are also in this class, see \textit{between} (80) and \textit{far} (100) in \autoref{tab:seed_probs}.

\begin{table*}[t]
\centering
\scalebox{.83}{
\begin{tabular}{l l}
\\
\textbf{Class} (\textbf{\#patterns}) & \textbf{Spatial expression counts} 
\\\midrule
Directional (95) 
& 
\tabul{in (20), from (17), into (9), to (8), on (8), away from (7), towards (7), out of (4), back (3),\\ 
through (3), across (2), at (2), outside (2), opposite (1), part of (1), by (1)}
\\\midrule
Argument orientation (67) 
& 
\tabul{in (21), at (10), from (9), away from (4), out of (4), near (3), with (3), inside (3), on (2), \\
under (2), through (1), opposite (1), towards (1), far from (1), on top of (1), behind (1)}
\\\midrule
Projective (70) 
& 
\tabul{behind (16), between (11), in front of (10), below (6), above (6), under (6), on top of (5), \\
front of (3), opposite (2), to the right of (2), on (2), to the left of (1)}
\\\midrule
Non-projective (48)
& between (22), in (9), far from (5), close to (4), outside (3), on top of (2), on (2), opposite (1) 
\\\midrule
\end{tabular}
}
\caption{The spatial expressions and their counts per entailment class in the SpaceNLI patterns
}
\label{tab:exp_count}
\end{table*}

\section{Dataset construction}\label{sec:data}

\subsection{Pattern construction}
\label{ssec:pattern}

Patterns are labeled NLI problems with NPs replaced by variables as illustrated in \autoref{fig:intro}.
The NLI patterns are obtained from the seed NLI problems.
To collect the latter, we extracted the initial 56 problems from \citet{zwarts2000vector} and \citet{nam:1995}, where a majority of the problems were labeled as entailment due to obvious biases in the semantic literature towards licensing entailment.
To create a representative and challenging NLI dataset for machine learning, we applied several \emph{revision phases} to the problems: introducing new problems that either cover new semantic aspects of spatial expression or serve as a perturbed version of an existing problem.

In the initial revision phase, four annotators divided the extracted problems and created slightly modified versions of them with an inference label different from the original.%
\footnote{
The annotators for the pattern construction consist of the authors of the paper, two linguist students, and one AI student.
The guideline for creating inference problems can be found in the supplementary material.
}
This was motivated by the current trends in the literature on adversarial, stress, and debiased datasets (\citealt{naik-etal-2018-stress,ribeiro-etal-2020-beyond,kaushik2020learning,gardner-etal-2020-evaluating}, \textit{inter alia}).
For example, (16) is a perturbed example of (15). 
Where possible, NLI problems of a new type were also created using the similar spatial expressions found in the extracted problems.

To validate the resulting pool of NLI problems (in total 162), following \citep{zhang-etal-2017-ordinal}, they were labeled on a 5-point Likert scale by three annotators.%
\footnote{
The question was to what extent the hypothesis sentence is true, given that the premises are true, with choices: \emph{definitely false}, \emph{most likely false}, \emph{unknown}, \emph{most likely true}, \emph{definitely true}.
We used two additional choices, \emph{difficult} (unable to annotate due to the complex reasoning it requires) and \emph{skip} (presence of an ungrammatical or nonsensical sentence).  
We used the brat annotation tool \citep{stenetorp-etal-2012-brat} for labeling.
The annotation guideline is included in the supplementary material.
}
After collecting the 5-point annotations, for each annotator, we picked a mapping of 5-point to 3-point that maximizes the inter-annotator agreement (avg. Cohen's $\kappa=.71$).     
The problems without majority labels were discarded and 111 problems remained.

To better balance the inference labels and increase the coverage of spatial expressions, a second revision phase was carried out on the remaining problems.
In several cases, problems with low annotator agreement were revised, e.g., changing the tense where it caused confusion or replacing a preposition with a weaker version (\textit{at}$\mapsto$\textit{near}). 
All the new and revised problems (in total 63) were validated based on three samples:
each problem was manually converted into a pattern by replacing NPs with variables, and three random NLI samples per pattern were generated (see \autoref{ssec:sample} for details), which were subsequently validated by three annotators.

Finally, a third revision phase was carried out on the remaining problems to additionally decrease the overall and spatial type-specific label imbalance.
The collected problems (in total 160) were treated as a seed by converting them into NLI patterns to generate a large amount of sample NLI problems from them.
To illustrate the coverage of spatial expressions in the collected patterns, \autoref{tab:exp_count} gives the complete list of spatial expressions for each entailment class.

\subsection{Sample generation}
\label{ssec:sample}

We manually created NLI patterns from the initially collected NLI problems (\autoref{ssec:pattern}) by replacing NPs with placeholders and specifying selection restrictions for them imposed by the verbs, spatial expressions, and gold inference labels (see \autoref{fig:intro}).
The selection restrictions imposed by spatial expressions are subtle and can affect gold labels or the naturalness of sentences.
For example, if the figure is much larger than the ground, it can make the sentence infelicitous:
\textit{the apple on the fridge} and \textit{the apple near the fridge} are preferred to \textit{the fridge under the apple} and \textit{the fridge near the apple}.
Inferences driven by proximity-related spatial expressions are sensitive to the size of the objects.
For instance, based on our conducted validations, \textit{Cindi is opposite to the cat} is more likely to be neutral to \textit{Cindi is far from the cat}, but \textit{the school is opposite to the house} is more likely to contradict \textit{the school is far from the house}.

To meet selection restrictions and allow relative diversity of NPs in the generated samples, we defined a mini world with a domain containing 171 entities corresponding to common and proper nouns.
The entities are organized in a taxonomy with 20 subclasses covering general types of entities (e.g., person, animal, vehicle), the projections of an argument in certain argument structures (e.g., enter in $X$, be in $X$, throw $X$), compatibility with projective spatial expressions, and size categories (S for entities comparable to small objects like book and cat, M to persons, and L to vehicles).
Binary and ternary relations are defined based on the set unions of the products of entity sets and subclasses.

To automatize the sampling of sound NLI problems from the patterns, we formatted the mini world in YAML and NLI patterns in XML.
We implemented a procedure that samples problems from the patterns by filling in NP placeholders with definite NPs from the mini world and respecting the pattern-specific selection restrictions.
For sanity checking, the procedure verifies that it can generate corresponding seed NLI problems for each pattern.

To measure how faithfully the inference labels are transferred from seed and pattern NLI problems to the corresponding NLI samples, we used sampled problems in the second phase of validation when validating new NLI problems (see \autoref{ssec:pattern}). 
The results showed that 79\% of samples were unanimously labeled with the original label.
After filtering out patterns with a relatively low agreement, this ratio increased to 97\% for the samples generated from the validated patterns.

The NLI problems sampled from the same pattern or related patterns are string-wise very close to each other, sometimes differing only in terms of occurrences of a single NP.
Regardless of this similarity, we expect such problems to pose a challenge for NLI systems based on large language models (LLMs) as it has been shown that their predictions can be sensitive to a single-word substitution \citep{glockner-etal-2018-breaking,gururangan-etal-2018-annotation}.
In addition to NPs, one could have allowed the replacement of other phrases in the NLI patterns, but this would have significantly complicated the definition of the mini world and generation of natural and sound NLI samples.     

\begin{table}[t!]
\centering
\scalebox{.9}{\begin{tabular}{@{}lrrrr@{}}
\toprule
Property\kern-0mm &     E \% &     N \% &     C \% &    All \% (\#)\kern5mm \\
\midrule
Dir  &  39.6 &  35.4 &  25.0 &    30.0\phantom{0} ~ (9600) \\
NonP &  25.0 &  41.7 &  33.3 &    22.5\phantom{0} ~ (7200) \\
Proj &  29.4 &  26.5 &  44.1 &    21.2\phantom{0} ~ (6800) \\
ArgO &  47.6 &  28.6 &  23.8 &    26.2\phantom{0} ~ (8400) \\
\midrule
$+$ neg  &  48.0 &  28.0 &  24.0 &    15.6\phantom{0} ~ (5000) \\
\midrule
1prem   &  41.8 &  26.5 &  31.6 &   61.3 ~ (19600) \\
2prem   &  25.0 &  42.9 &  32.1 &   35.0 ~ (11200) \\
3prem   &  50.0 &  50.0 &   0.0 &    3.8 ~ \phantom{0}(1200) \\
\midrule
All  &  36.2 &  33.1 &  30.6 &  100.0 ~ (32000) \\
\bottomrule
\end{tabular}}
\caption{Statistics of several properties of the sampled NLI dataset.
The statistics also apply to the collection of NLI patterns as the samples are evenly distributed over the patterns.
The properties consist of the spatial inference types, whether including negation, and the number of premises.}
\label{tab:data_stats}
\end{table}

\begin{table}[th!]
\centering
\scalebox{.87}{\begin{tabular}{@{}l@{}crrrr}
\toprule
\multirow{2}{*}{\tabuc{LLM-based\\NLI models}}
& \multirow{2}{*}{\tabuc{Training\\data}}
& \multirow{2}{*}{\tabuc{\textsc{snli}\\[-2mm]+\\[-2mm]\textsc{mnli}}} &  \multicolumn{3}{c}{SpaceNLI}\\
& & & Acc &  \kern-2mmPA$_{0.95}$\kern-2mm &  \kern-3mmPA$_{1.0}$\kern-2mm \\
\midrule
\tabul{\footnotesize DeBERTaV3-L\#1\\[-2mm]
\scriptsize{\href{https://huggingface.co/Joelzhang/deberta-v3-large-snli_mnli_fever_anli_R1_R2_R3-nli}{Joelzhang/deberta-v3\dots}}} 
& \textsc{smfa}  & \textbf{91.8} &   59.6 &  47.5 & \textbf{37.5} 
\\
\tabul{\footnotesize ALBERT-XXLv2\\[-2mm]
\scriptsize{\href{https://huggingface.co/ynie/albert-xxlarge-v2-snli_mnli_fever_anli_R1_R2_R3-nli}{ynie/albert-xxlarge-v2\dots}}} 
& \textsc{smfa}  & 90.8 &   57.8 &  \textbf{48.1} &  36.2 
\\
\tabul{\footnotesize DeBERTa-L\\[-2mm]
\scriptsize{\citet{he2021deberta}}}                       
& \textsc{m}     & 90.7 &   54.1 &  42.5 &  36.2 
\\
\tabul{\footnotesize RoBERTa-L\\[-2mm]
\scriptsize{\citet{nie-etal-2020-adversarial}}} 
& \textsc{smfa}  & 90.6 &   55.6 &  40.0 &  31.9 
\\
\tabul{\footnotesize BART-L\\[-2mm]
\scriptsize{\href{https://huggingface.co/ynie/bart-large-snli_mnli_fever_anli_R1_R2_R3-nli}{ynie/bart-large-snli\_mnli\dots}}}  
& \textsc{smfa}  & 90.4 &   55.4 &  39.4 &  29.4 
\\
\tabul{\footnotesize DeBERTaV3-L\#2\\[-2mm]
\scriptsize{\citet{OSF_data_scarcity_2022}}} 
& \textsc{mfalw} & 90.3 &   \textbf{66.5} &  44.4 &  33.8 
\\
\tabul{\footnotesize XLNet-L-cased\\[-2mm]
\scriptsize{\citet{nie-etal-2020-adversarial}}} 
& \textsc{smfa}  & 90.3 &   55.8 &  42.5 &  30.0 \\
\bottomrule
\end{tabular}}
\caption{Performance of SOTA NLI systems on SpaceNLI. 
\textsc{snli+mnli} shows the average score on these datasets.
Training data names are denoted with the initial letters: \textbf{S}NLI, \textbf{M}NLI, \textbf{A}NLI, \textbf{F}ever-NLI, \textbf{W}ANLI, and \textbf{L}ingNLI. 
The best system per problem accuracy on SpaceNLI, DeBERTaV3-L$_{\text{MFALW}}$ (with $\Delta\geq6.9\%$), doesn't turn out to be the best at the consistency threshold $\geq0.95$.
See the extended version of the table in \autoref{sec:appendix}.
}
\label{tab:sota_nli}
\end{table}

\section{Experiments} \label{sec:exp}

\subsection{Sample dataset} \label{ssec:generate_data}

We uniformly generated a spatial dataset of 32,000 NLI samples from 160 NLI patterns, i.e., 200 samples per pattern.
We used the mini world as described in \autoref{ssec:sample}.
The dataset statistics are given in \autoref{tab:data_stats}.
The inference labels are relatively balanced: each label being represented by at least 30\% of the problems. 
Each spatial inference type counts at least 20\% of the overall problems and 23\% of label-specific problems. 
In contrast to the common biases in NLI datasets, a majority of the problems with negation are labeled as entailment, not contradiction.
This is due to perturbed problems introduced in the revision phases (\autoref{ssec:pattern}). 
Around 39\% of problems have multiple premises, where three-premised problems occur only in the directional problems, the argument orientation problems contain only single-premised problems, and most of the multi-premised problems are in the non-projective problems.
We refer to the generated dataset as SpaceNLI and use it in subsequent experiments.%
\footnote{We make the collection of the patterns, the generation code, and the sample dataset publicly available upon the acceptance of the paper.
}
\subsection{Evaluating SOTA NLI systems}

\subsubsection{Standard accuracy}

We selected NLI models that have results comparable to the state of the art in NLI and evaluate them on SpaceNLI.
The models were chosen based on their availability, tractable size, and high average accuracy ($>90\%$) on the SNLI \citep{bowman-EtAl:2015:EMNLP} and MNLI \citep{multinli-18} datasets (see \autoref{tab:sota_nli}).
The models are based on various large language models (LLMs) like DeBERTaV3 \citep{he2023debertav}, BART \citep{lewis-etal-2020-bart}, ALBERT \citep{lan2020albert}, XLNet \citep{yang2020xlnet}, etc. (see \autoref{tab:sota_nli}).
The LLMs are fine-tuned on several NLI train datasets: SNLI, MNLI, FEVER-NLI \citep{nie2019combining}, ANLI \citep{nie-etal-2020-adversarial}, LingNLI \citep{parrish-etal-2021-putting-linguist}, WANLI \citep{liu-etal-2022-wanli}.
We use the models from the HuggungFace model hub%
\footnote{\url{https://huggingface.co/models}}
and provide them with the corresponding hub names in \autoref{tab:sota_nli}. 

The results in \autoref{tab:sota_nli} show that \mbox{DeBERTaV3-L\#2} trained on a large collection of training datasets (885K problems in total) generalizes best on the spatial reasoning (66.5\%), achieving a substantial improvement ($\geq 6.9\%$) over the other models.%
\footnote{The second best, \mbox{DeBERTaV3-L\#1}, is based on the same LLM fine-tuned on a different combination of NLI datasets. 
Note that \citet{OSF_data_scarcity_2022} deliberately removed SNLI from the training set as it negatively affected the accuracy of the model in their experiments.
}

\subsubsection{Consistency \& pattern accuracy}

To evaluate the models on the consistency of their predictions for NLI problems from the same pattern, we define the pattern accuracy (PA) score and its curve.  
The PA curve records the PA score of a model for each consistency threshold. 
Informally, the PA score with a consistency threshold $t$ is a ratio of NLI patterns for which model gets at least $t$ portion of the samples generated from them.
For example, the PA of 50\% with a threshold 90\% means that there are a half of the NLI patterns such that for each pattern a model is able to correctly classify at least 90\% of its sample problems.    
The formal definition of the PA with a threshold $t$ is:

\medskip
\centerline{\scalebox{.95}{$\displaystyle
P\!A_t(\hat{Y}, \mathbf{y}) = 
\frac{1}{N}\sum_{i=1}^N 
\bigg[ 
\frac{\sum_{k=1}^{M_i} \delta(\hat{y}_k^i =y^i)}{M_i} \geq t 
\bigg]
$}}
\medskip

\noindent where $\hat{Y} = (\hat{y}^i_k)_{1\leq i \leq N, 1\leq k \leq M_i}$ are predictions for $k^\text{th}$ sample of $i^\text{th}$ pattern, 
$N$ is the number of patterns, 
$M_i$ is the number of samples for $i^\text{th}$ pattern,
$\mathbf{y} = (y^i)_{1\leq i \leq N}$ gold labels of $i^\text{th}$ pattern, and $\delta$ is the Kronecker delta.

\begin{figure}[t!]
\includegraphics[clip, trim=4mm 2.5mm 4mm 4mm, width=.48\textwidth]{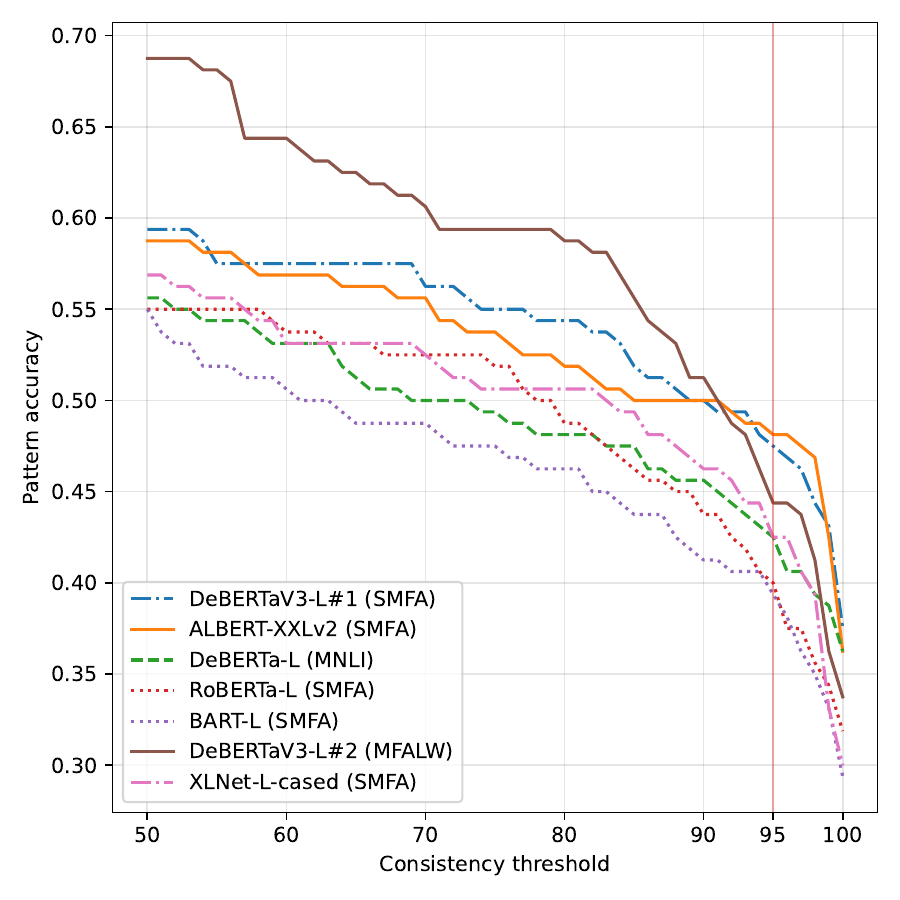}
\caption{Pattern accuracy curves of the NLI models from \autoref{tab:sota_nli}.
The first half, which corresponds to the scores allowing solving less than half of the samples per pattern, is omitted (see \autoref{sec:appendix} for the complete curves). 
}
\label{fig:const-curves}
\end{figure}

While DeBERTaV3-L\#2 gets the best score on the SpaceNLI problems, based on the PA scores in \autoref{tab:sota_nli}, it shows high consistency ($P\!A_{0.95}$ or $P\!A_{1.0}$) in fewer NLI patterns than the other two  competing models, DeBERTaV3-L\#1 and ALBERT-XXLv2.
PA curves of the NLI models provide a closer look at this contrast (see \autoref{fig:const-curves}). 
While the curve of DeBERTaV3-L\#2 outperforms other models by a margin, it is noteworthy that it does this by classifying sample problems of the patterns which it can hardly solve half of the time (this is visible in the complete curves in \autoref{sec:appendix}). 
It drastically decreases after 95\% of consistency while ALBERT-XXLv2 and DeBERTAV2-L\#1 maintain very high consistency for $>47$\% of NLI patterns.
This demonstrates that a high-performing model is not necessarily the most consistent across patterns.

RoBERTa-L and BART-L obtain similar accuracy scores, but RoBERTa-L is more consistent in more NLI patterns than BART-L while the latter gets slightly more NLI problems for inconsistently predicted patterns.
The complete curves in \autoref{sec:appendix} shows how the curves swap places after the consistency threshold of 50.
This shows that the standard accuracy (i.e., based on NLI problem samples) can blur the fine distinction in consistency between the models.   

The dispersion of the curves at the lowest end of the consistency threshold is twice larger than at the highest end.
This shows that the model predictions more diverge in coverage of patterns than in consistency per pattern.  
In other words, the contrast confirms the sensitivity of the models towards the inference-preserving word substitutions.

\subsubsection{Few-shot learning experiments}

\begin{figure}[t!]
\mbox{\includegraphics[clip, trim=3mm 0mm 10mm 15mm, width=.50\textwidth]{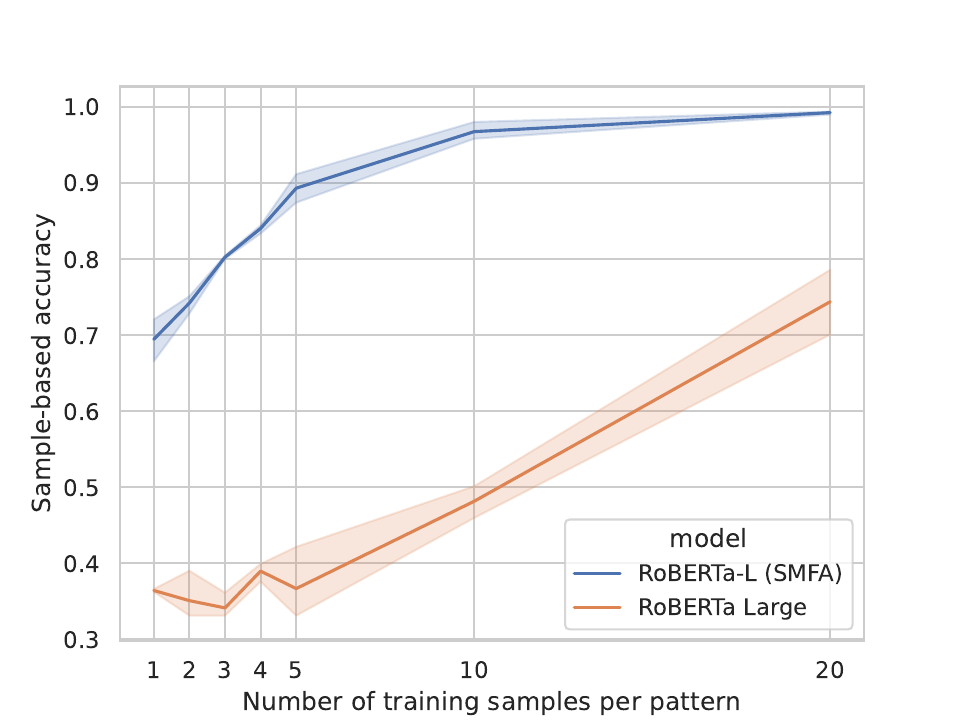}}
\caption{Average of three runs for each few-shot finetuning experiment.
RoBERTa-L (SMFA, \citealt{nie-etal-2020-adversarial}) is already finetuned on several large NLI datasets while RoBERTa Large \citep{liu2019roberta} is a pretrained language model without any previous training on NLI.   
}
\label{fig:few-shot}
\end{figure}

We measured the difficulty of the SpaceNLI problems in terms of few-shot learning experiments.
We used 100 samples per pattern as a test set while other 100 samples per pattern were used for drawing a few samples for each pattern.
In this way, the patterns are fully shared between the training and test sets, but no sample NLI problem is in both sets. 
For each number of shots, we carried out the sample drawing process three times.
We used two NLI models: a high performing NLI model RoBERTa-L$_\text{SMFA}$ from \citet{nie-etal-2020-adversarial} and a \emph{vanilla} NLI model based on the large RoBERTa pretrained language model \citep{liu2019roberta}.  
The results of the few-shot experiments are in \autoref{fig:few-shot}.

Finetuning RoBERTa-L$_\text{SMFA}$ on a single sample of each pattern increases the sample-based accuracy on the test set by 14\%.
Each additional sample further boosts the model's accuracy.
The almost perfect accuracy (>99\%) is reached when 20 samples per pattern are seen during the finetuning.
The results show that the lexical variability poses a challenge to the high-performing NLI model as it needs to be finetuned on at least five samples for every pattern of the test set to achieve a high score.

The challenge coming from the lexical variability and the SpaceNLI patterns is further emphasized by the relatively low results of RoBERTa Large.
Even after being finetuned on the 20 samples of each NLI pattern, the model is still far from the high performance on unseen samples (but seen patterns).
The relatively low results can be also partially attributed to the low ratio between the number of training samples and the large number of the model's trainable parameters.

\begin{figure*}[t!]
\mbox{\includegraphics[clip, trim=4mm 4mm 4mm 4mm, width=\textwidth]{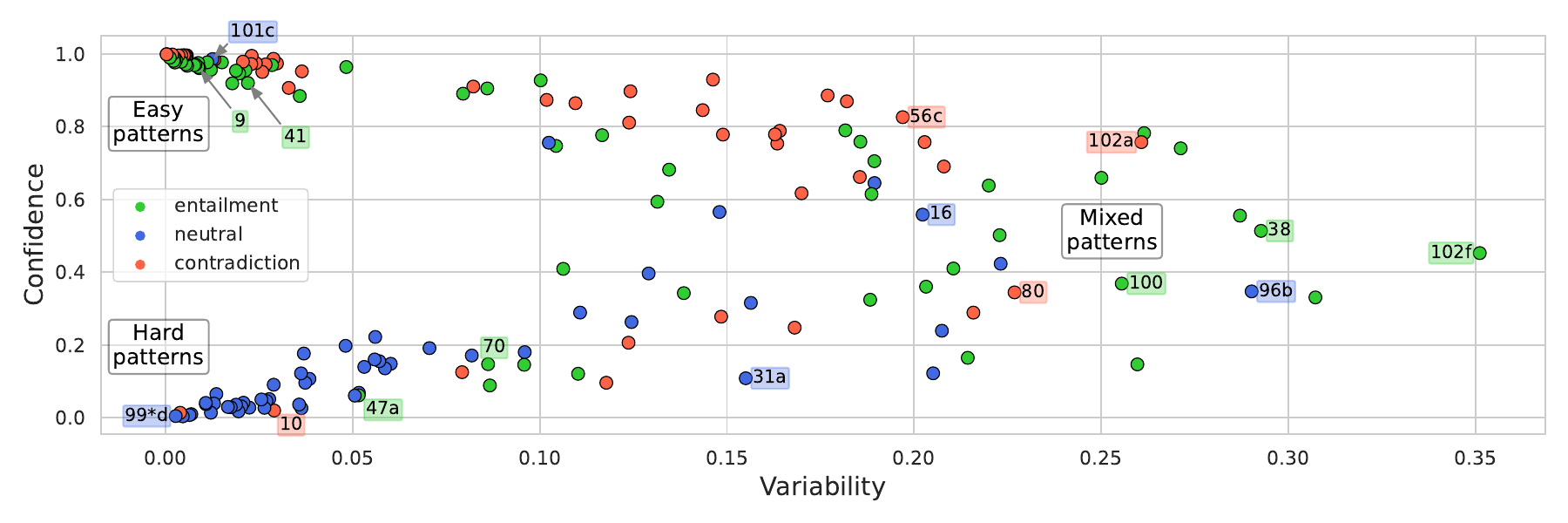}}
\caption{Prediction cartography of \texttt{RoBERTa-large} from \cite{nie-etal-2020-adversarial}. 
NLI patterns are characterized with \emph{confidence} and \emph{variability}: the mean and the standard deviation of probabilities assigned by the model to the true labels of the sample NLI problems.
IDs mark NLI patterns from \autoref{fig:intro} and \autoref{tab:seed_probs}.
}
\label{fig:anli-catrography}
\end{figure*}

\section{Analysis} \label{sssec:analysis}

\begin{figure}[t!]
\mbox{\includegraphics[clip, trim=7mm 5mm 10mm 17mm, width=.51\textwidth]{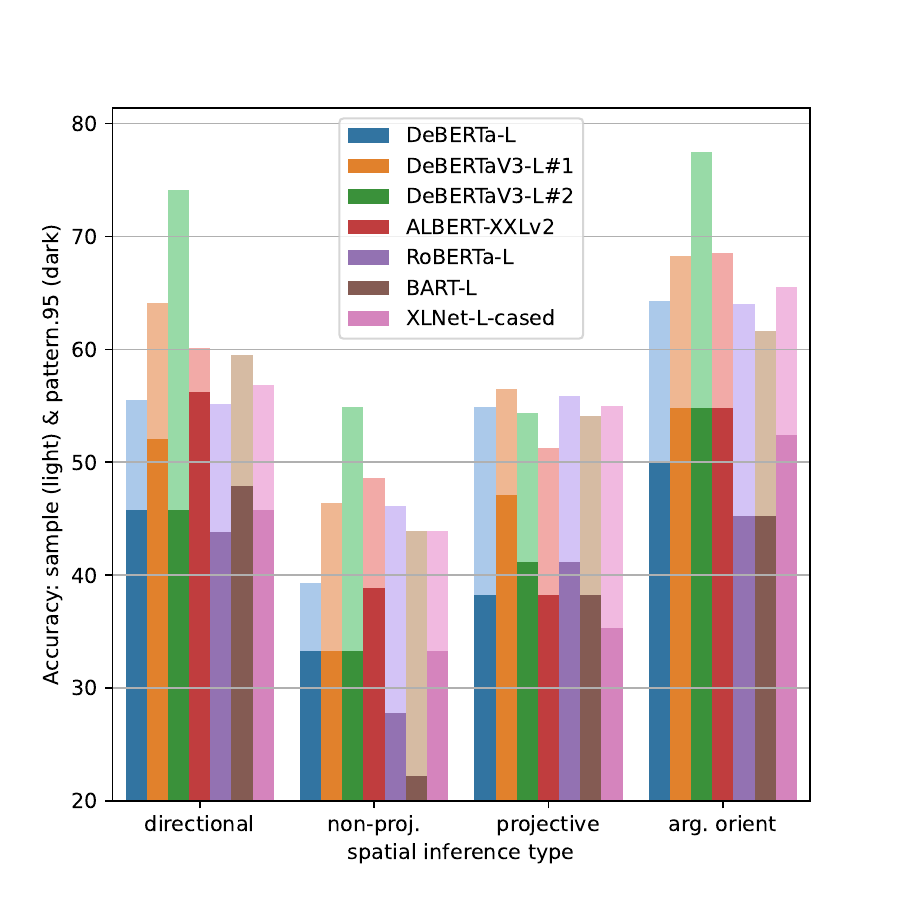}}
\caption{Sample-based (in light shades) and $P\!A_{0.95}$ (in dark shades) accuracy scores of the models per spatial inference type.
}
\label{fig:space-types}
\end{figure}

To find out what type of inferences the models find challenging, we analyze the models' performance per inference type. 
\autoref{fig:space-types} shows the sample- and pattern-based accuracy scores of the models per spatial inference types as defined in  \autoref{ssec:inf_types}.
The model ranking based on the sample accuracy varies across the inference types.
For instance, the best model, DeBERTaV3-L\#2, remains at the top of the rankings for all inference types with quite a margin except for the projective type.
On average, non-projective spatial inferences are the most challenging for the models.
The easiest of the types is argument orientation, the type that is closest to the PP attachment task.
For the other inference types, projective inferences are harder than directional ones.
The apparent distinction in the scores between the inference types is also preserved for the $P\!A_{0.95}$ score (shown with the dark bars in \autoref{fig:space-types}).
The fine-grained analysis additionally shows that the best model, DeBERTaV3-L\#2, suffers least in terms of consistency on the projective inferences while its performance on this inference type is not among the best.

Based on the results in \autoref{fig:space-types}, the non-projective NLI patterns and samples are the most challenging for the SOTA models.
When looking closer at the set of non-projective problems, it turns out that it contains a high number of problems (46\%) with the spatial expression ``between`` (as shown in \autoref{tab:exp_count}), and these problems are specially challenging due to complex semantics of ``between''.
The average accuracy of the models on such NLI samples is 41.6\%.
This is lower than the average sample-based accuracy (46.1\%) on entire SpaceNLI and much lower than the average sample-based accuracy (54.1\%) on the other part of the non-projective samples.

We further zoom in on the NLI patterns and measure a model's probabilistic predictions for the patterns.
Namely, following \citet{swayamdipta-etal-2020-dataset}, we measure a model's confidence and variability.
Originally the dataset cartography \citep{swayamdipta-etal-2020-dataset} was used to analyze the training dynamics of a model across the epochs and identify training samples that are easy or difficult for learning.
In contrast, we use dataset cartography for analyzing evaluation dynamics across patterns and identifying easy and hard ones.%
\footnote{Put differently, iterative classification of the same training sample across epochs, is replaced with the classification of the same NLI pattern based on its samples.
}

\autoref{fig:anli-catrography} illustrates the pattern-based evaluation dynamics of RoBERTa-L \cite{nie-etal-2020-adversarial}, an average model based on the evaluations.
For instance, NLI pattern (102f) happens to have one of the most variable samples according to the model predictions: the mean and the standard deviation of the probabilities the model assigns to the entailment class of the samples of (102f) are 0.45 and 0.35, respectively.    

\medskip
\begin{tabular}{r}
(\texttt{102f}) ~~~ NP$_1$ has hidden NP$_2$ behind NP$_3$.\\ 
\texttt{entailment} ~~~ NP$_2$ is not in NP$_3$.
\end{tabular}
\medskip

\noindent The evaluation cartography shows that the predictions vary mostly for entailment patterns (in green).
Most of the hard patterns are neutral ones (in blue) and vice versa.
Contradiction patterns (in red) tend to be easy with some variability.

\section{Related work}\label{sec:rel_work}

Several works have automatically sampled NLI problems from curated patterns/templates.
\citet{jeretic-etal-2020-natural} generated the implicature and presupposition diagnostic dataset IMPPRES from pre-defined templates.  
\citet{mccoy-etal-2019-right} constructed the HANS dataset by designing templates of NLI problems that support or refute certain inference heuristics, which were later used to generate NLI problems.  
\citet{richardson_sem_frag:2020} used the template language from \citet{salvatore-etal-2019-logical} to produce NLI problems involving negation, Boolean connectives, quantifiers, cardinals, conditionals, and comparatives.
These works all use restricted vocabulary while generating samples from the patterns. 

With its pattern-based construction and restricted vocabulary, SpaceNLI comes close to the IMPPRES \citep{jeretic-etal-2020-natural} and HANS \citep{mccoy-etal-2019-right} datasets.
Unlike these datasets, SpaceNLI involves multiple-premised problems and puts more emphasis on satisfying selection restrictions to prevent nonsensical sentences.

Based on the nature of NLI problems, SpaceNLI resembles FraCaS \citep{fracas} as both contain inference problems often found in textbooks on formal semantics.
Unlike FraCaS, the inference labels of patterns in SpaceNLI are quite balanced and the number of spatial NLI patterns is twice the size of the largest section in FraCaS.   

There have been attempts to identify semantic phenomena in existing NLI datasets, including aspects of spatial reasoning.
By looking up certain keywords, \citet{kim-etal-2019-probing} automatically detect NLI problems in MultiNLI \citep{multinli-18} that might contain spatial expressions.
They create a mutated sample from the original NLI problem by negating the sentence with the potential spatial expression. 
\citet{joshi-etal-2020-taxinli} annotate MultiNLI problems based on the semantic aspects required by the inference label.
Their taxonomic categories include the spatial subcategory, grouped with the relational, temporal, causal, and co-reference subcategories.

The problems in SpaceNLI are substantially diverse from a semantic perspective than the MultiNLI problems that were identified by \citet{kim-etal-2019-probing} and \citet{joshi-etal-2020-taxinli}.
The MultiNLI dataset is crowd-elicited and doesn't have problems with sufficient depth in spatial reasoning. 

\section{Conclusion}

To the best of our knowledge, we have created the first spatial inference dataset that involves diverse spatial inference types.
The structure and the evaluation protocol are unique as we focus on performance on the NLI patterns and consistency across the samples in the pattern, instead of focusing on mere quantitative accuracy based on the NLI problems/samples. 
The evaluation protocol tests models whether they can consistently recognize inference patterns while generalizing over \emph{irrelevant} lexical substitutions.
The more consistent a model is in its predictions, the less unexpected its behavior becomes.

The SOTA NLI models show moderate generalization capacity on spatial problems.
While the top-performing model gets the highest overall accuracy, it is ranked third when it comes to the consistency of predictions inside the patterns: predicting at least 95\% of the samples per pattern.

The introduced pattern accuracy (PA) curves provide a more fine-grained distinction between the models: the models with comparable standard accuracy scores might substantially differ in the consistency of their predictions.
Overall the performance of models drops ca. 10\% when raising the consistency threshold to 95\%.  
This illustrates that the predictions of the SOTA models are sensitive to lexical replacements that have no effect on the semantics of the inference.   

The evaluation results revealed that the most challenging inference type is associated with non-projective locatives mainly due to the complex semantics of ``between'' while the argument orientation type is the easiest.
The latter is somewhat expected as the problems in the argument orientation type are close to the task of PP attachment which LLMs are expected to be good at.

\section*{Acknowledgments}

This work was funded by the European Research Council (ERC) under the European Union's Horizon 2020 research and innovation programme (grant agreement No 742204). 
We would like to acknowledge the help from three student assistants with the data annotation and thank the anonymous reviewers for their helpful comments.

\bibliographystyle{acl_natbib}
\bibliography{main}

\newpage
\appendix

\begin{table*}[th!]
\centering
\scalebox{.87}{\begin{tabular}{@{}l@{~}rrrrrrrrrr@{}}
\toprule
\multirow{2}{*}{\tabul{LLM-based NLI models \small{(train data)}\\[-1mm]
\scriptsize{model names from Huggingface hub}} } 
&  \multirow{2}{*}{\textsc{snli}} 
& \multirow{2}{*}{\textsc{m}$_{\text{m}}$} & \multirow{2}{*}{\textsc{m}$_{\text{mm}}$}
& \multirow{2}{*}{\textsc{s+m}} &  \multicolumn{6}{c}{SpaceNLI (accuracy \& $\geq$\,consistency score)}\\
& & & & & Acc & \footnotesize $\geq0.5$ &  \footnotesize $\geq0.67$ &   \footnotesize $\geq0.9$ &  \footnotesize $\geq0.95$ &   \footnotesize $=1.0$ \\
\midrule
\tabul{\footnotesize DeBERTaV3-L\#1 (SMFA)\\[-2mm]
\tiny{Joelzhang/deberta-v3-large-snli\_mnli\_fever\_anli...}} 
&       \textbf{92.9} &        \textbf{91.4} &         \textbf{91.2} &   \textbf{91.8} &   59.6 &  59.4 &  57.5 &  50.0 &  47.5 &  \textbf{37.5} 
\\
\tabul{\footnotesize ALBERT-XXLv2 (SMFA)\\[-2mm]
\tiny{ynie/albert-xxlarge-v2-snli\_mnli\_fever\_anli\_R1\_...}} 
&       91.9 &        90.2 &         90.2 &   90.8 &   57.8 &  58.8 &  56.2 &  50.0 &  \textbf{48.1} &  36.2 
\\
\tabul{\footnotesize DeBERTa-L (MNLI) \citep{he2021deberta}\\[-2mm]
\tiny{microsoft/deberta-large-mnli}}                       
&       89.6 &        91.3 &         91.1 &   90.7 &   54.1 &  55.6 &  50.6 &  45.6 &  42.5 &  36.2 
\\
\tabul{\footnotesize RoBERTa-L (SMFA) \citep{nie-etal-2020-adversarial}\\[-2mm]
\tiny{ynie/roberta-large-snli\_mnli\_fever\_anli\_R1\_R2\_R...}} 
&       91.8 &        89.9 &         90.0 &   90.6 &   55.6 &  55.0 &  52.5 &  43.8 &  40.0 &  31.9 
\\
\tabul{\footnotesize BART-L (SMFA)\\[-2mm]
\tiny{ynie/bart-large-snli\_mnli\_fever\_anli\_R1\_R2\_R3-nli}}  
&       92.0 &        89.4 &         89.6 &   90.4 &   55.4 &  55.0 &  48.8 &  41.2 &  39.4 &  29.4 
\\
\tabul{\footnotesize DeBERTaV3-L\#2 (MFALW) \citep{OSF_data_scarcity_2022}\\[-2mm]
\tiny{MoritzLaurer/DeBERTa-v3-large-mnli-fever-anli-l...}} 
&       89.0 &        91.2 &         90.8 &   90.3 &   \textbf{66.5} &  \textbf{68.8} &  \textbf{61.9} &  \textbf{51.2} &  44.4 &  33.8 
\\
\tabul{\footnotesize XLNet-L-cased (SMFA) \citep{nie-etal-2020-adversarial}\\[-2mm]
\tiny{ynie/xlnet-large-cased-snli\_mnli\_fever\_anli\_R1\_...}} &       91.7 &        89.8 &         89.5 &   90.3 &   55.8 &  56.9 &  53.1 &  46.2 &  42.5 &  30.0 \\
\bottomrule
\end{tabular}}
\caption{Performance of NLI models on SpaceNLI and common NLI benchmarks: SNLI-test, MNLI-val-matched, and MNLI-val-mismatched. 
S+M shows the average of the three accuracy scores.
Training data names are denoted with the initial letters: \textbf{S}NLI, \textbf{M}NLI, \textbf{A}NLI, \textbf{F}ever-NLI, \textbf{W}ANLI, and \textbf{L}ingNLI. 
The best model per problem accuracy on SpaceNLI, DeBERTaV3-L$_{\text{MFALW}}$ (with $\Delta\geq6.9\%$), doesn't turn out to be the best at the consistency threshold $\geq0.95$.   
}
\label{tab:sota_nli_exp}
\end{table*}

\begin{figure*}[t!]
\mbox{\includegraphics[clip, trim=4mm 4mm 4mm 4mm, width=\textwidth]{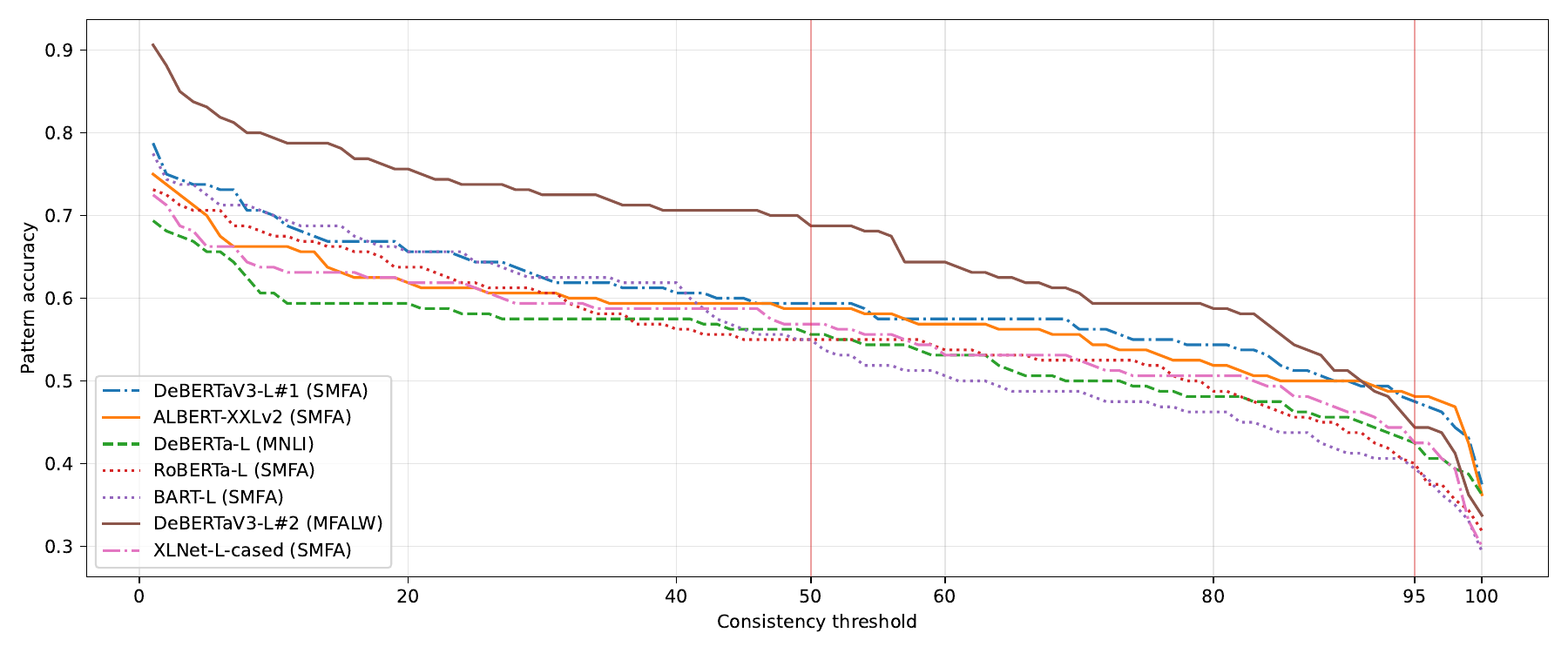}}
\caption{Pattern accuracy curves of the NLI models from \autoref{tab:sota_nli}.
The area under the curve represents a standard NLI problem-based accuracy.
}
\label{fig:PA_curves_long}
\end{figure*}

\section{Results}
\label{sec:appendix}

\autoref{tab:sota_nli_exp} represents the extended version of \autoref{tab:sota_nli}.
Note that the area under the curve corresponds to the standard accuracy based on the NLI problems.

\end{document}